\documentclass{article} 
\usepackage{iclr2027_conference,times}

\usepackage{amsmath,amsfonts,bm}

\def\eqref#1{equation~\ref{#1}}
\def\Eqref#1{Equation~\ref{#1}}

\def\1{\bm{1}}

\DeclareMathAlphabet{\mathsfit}{\encodingdefault}{\sfdefault}{m}{sl}
\SetMathAlphabet{\mathsfit}{bold}{\encodingdefault}{\sfdefault}{bx}{n}

\renewcommand{\eqref}[1]{(\ref{#1})}
\renewcommand{\Eqref}[1]{Eq.~\eqref{#1}}

\usepackage[hidelinks]{hyperref}
\usepackage{url}
\usepackage{booktabs}
\usepackage{multirow}
\usepackage{graphicx}
\usepackage{subcaption}
\usepackage{float}

\title{\texorpdfstring{Transform-Aligned Learned Features\\
for Lossy Point Cloud Attribute Compression}{Transform-Aligned Learned Features for Lossy Point Cloud Attribute Compression}}

\iclrfinalcopy
\author{%
Yueru Chen$^{1}$, Pengpeng Yu$^{1,2}$, Dingquan Li$^{1}$,
Wei Gao$^{3}$, Wei Zhang$^{1,4}$, and Fei Song$^{1}$ \\
$^{1}$Pengcheng Laboratory, Shenzhen, China \\
$^{2}$Sun Yat-sen University, Shenzhen, China \\
$^{3}$Peking University, Shenzhen, China \\
$^{4}$Xidian University, Xi'an, China
}

\begin{document}

\maketitle
\fancyhead{}
\renewcommand{\headrulewidth}{0pt}
\suppressfloats[t]

\begin{abstract}
Transform-based methods provide an effective framework for point cloud attribute
compression by representing attributes as transform coefficients.
Introducing learned spatial context into this framework requires mapping
spatial representations to the transform domain, but
this known basis change is often left for the network to learn implicitly.
We propose Transform-Aligned Learned Features (TALF) by applying the attribute transform to learned spatial
representations, explicitly aligning them with the coding targets.
Our analysis shows that the resulting features exactly represent the first-order prediction term of a smooth nonlinear model, with a bounded Taylor remainder.
We integrate TALF into a transform-based attribute codec with explicit coefficient
prediction and conditional residual entropy modeling under a unified
coefficient-domain rate--distortion objective, while retaining explicit
quantization-step control. Extensive experiments across three
benchmark datasets and multiple transform bases demonstrate that TALF
improves rate--distortion performance over conventional and learned baselines.
\end{abstract}

\section{Introduction}\label{sec:introduction}

Point cloud attributes such as reflectance and color carry information
needed for driving-scene perception and immersive 3D applications~\citep{graziosi2020overview,wang2025lidarsurvey}.
Practical attribute codecs benefit from two complementary capabilities:
expressive probability modeling for efficient compression~\citep{peng2025generalized,chen2025hapcac,3cacwang2023lossless} and an explicit,
interpretable quantization step (QS) for predictable rate--distortion control~\citep{wei2025qpc,zhang2024contentaware}.
Traditional transform codecs provide a well-defined transform--quantization
pipeline with reliable QS-based control, but rely on hand-crafted probability models with
limited contextual capacity. 
Learned approaches can capture richer dependencies~\citep{gao2025deepcompression,zhao2025lodpcac,nguyen2023lossless},
yet their latent representations can obscure the link
between quantization and attribute distortion~\citep{guo2025tscpcac,balle2021nonlinear,wang2022sparsepcac}. Effectively integrating these complementary strengths is therefore a promising direction for attribute compression.

Existing methods combine conventional attribute transforms with learned entropy modeling, as in 3DAC~\citep{fang2022_3dac}, or prediction, as in DeepRAHT~\citep{fu2026deepraht}.
However, the attribute transform defines the coding targets without itself aligning learned spatial representations. Consequently, the network is often left to learn this known basis change implicitly alongside coefficient prediction or probability modeling.

\begin{figure}[t]
\centering
\includegraphics[width=0.95\textwidth]{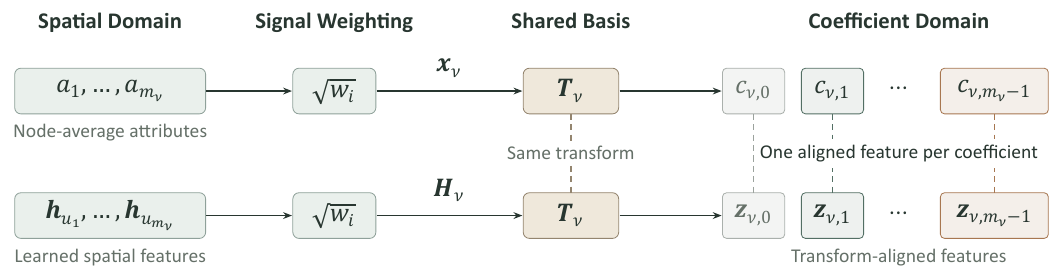}
\caption{Transform alignment within a hierarchical block.
Node-average attributes and learned spatial features undergo the same weighting and transform. Matching colors indicate coefficient--feature correspondence. Gray denotes the low-pass pair; only detail-aligned features enter the prediction and entropy-modeling heads.
}
\label{fig:talf_alignment}
\end{figure}

We propose Transform-Aligned Learned Features (TALF) to make this interface
explicit. As illustrated in Figure~\ref{fig:talf_alignment}, TALF applies the codec’s attribute transform to every learned feature channel, with the same weighting and node ordering as the attribute signals. 
The local smoothness and spatial correlation of point-cloud attributes~\citep{hu2022graphgeometric} motivate a local first-order approximation of a nonlinear predictor. Neighborhood-based linear prediction in conventional attribute codecs offers further practical support~\citep{do2023nonlinear}. We show that transform-aligned features
exactly represent this first-order prediction term and derive a bound on the
higher-order remainder, providing a structural basis for learned coefficient
modeling.

We instantiate TALF in an attribute codec with explicit
coefficient prediction and conditional residual entropy modeling.
We formulate the Laplace location parameter as an explicit coefficient prediction, jointly optimized through the entropy-rate and reconstruction-distortion terms. The resulting zero-centered residual is modeled conditionally.
Orthonormality preserves reconstruction distortion, enabling coefficient-domain rate–distortion learning that pairs each aligned feature with its coding target, rate, and distortion.
Analytical conversion between quantization units enables one model per
dataset to operate across the evaluated QS range. The alignment interface
accommodates different context encoders, compatible transform bases, and entropy-coding methods, with experiments covering a normalized graph basis and
RAHT. 
With learned entropy coding, TALF achieves BD-Rate savings of $20.93\%$ on Ford, $14.01\%$ on KITTI, and $9.71\%$ on ScanNet relative to our G-PCC anchor.

Our contributions are:
\begin{itemize}
    \item We introduce \textbf{transform-aligned learned features} through a
    parameter-free interface that applies the attribute transform to learned
    context features and establish its first-order alignment property with a bounded nonlinear remainder.
    \item We formulate a unified \textbf{coefficient-domain rate--distortion
    objective} that couples an explicit coefficient prediction with
    conditional entropy modeling of the resulting zero-centered residual.
    \item We integrate these components into a \textbf{QS-compatible lossy
    point cloud attribute coding framework}. Experiments on reflectance and
    color across datasets and transform bases demonstrate its rate--distortion benefits and the
    modularity of the alignment interface.
\end{itemize}

\section{Related Work}\label{sec:related_work}


\paragraph{Conventional Transform-Based Attribute Compression}
Traditional point cloud attribute compression exploits geometry--attribute
correlation through prediction and transform coding. 
The MPEG Geometry-based Point Cloud Compression (G-PCC) standard provides
prediction-, lifting-, and RAHT-based attribute coding
tools~\citep{schwarz2018emerging,mpeg2023gpccdescription}.
In particular, RAHT can be viewed as a geometry-adaptive variant of the Haar
wavelet transform, recursively applying occupancy-weighted two-point
transforms over the point-cloud octree hierarchy~\citep{de2016compression}. 
Graph-transform methods instead regard
attributes as signals over geometry-derived graphs~\citep{shao2017laplacian,xu2021predictive}. Early work established
blockwise graph Fourier coding for regular and sparse point
clouds~\citep{zhang2014point,cohen2016attribute}. RA-GFT incorporates multiresolution region masses
through a $Q$-normalized graph Laplacian~\citep{pavez2020ragft}, while SSGT
recursively constructs normalized-Laplacian subspace transforms~\citep{chen2020point}. 
Grounded in classical signal analysis, these methods provide well-defined transform–quantization pipelines. TALF builds on this structure by using the same transform to generate attribute coefficients and align the learned features used to model them.

\paragraph{Learning-Based Lossy Attribute Compression}
Existing learned methods broadly follow two routes according to the
representation being quantized and entropy-coded: learned latent
representations or explicit attribute transform coefficients.
The first route
includes Deep-PCAC, a geometry-conditioned point-based
autoencoder~\citep{sheng2022deeppcac}, and Unicorn, which employs universal
multiscale conditional coding for both lossy and lossless attribute
compression~\citep{wang2025unicorn}.
LVAC encodes learned latent representations in the form of RAHT coefficients for attribute reconstruction with local coordinate-based networks~\citep{isik2022lvac}. These methods offer expressive nonlinear modeling, but their reliance on learned representations makes rate–distortion control less interpretable.

The second route learns prediction or probability models for explicit
attribute transform coefficients. 3DAC retains RAHT and learns a context-adaptive probability model for arithmetic coding of the quantized transform coefficients~\citep{fang2022_3dac}.
DeepRAHT predicts node attributes through interpolation and learned
compensation, then applies RAHT to obtain coefficient predictions for
residual coding~\citep{fu2026deepraht}. It uses conventional zero run-length
coding rather than a learned conditional entropy model.
However, coefficient prediction or probability modeling does not itself ensure alignment between learned spatial representations and transform coefficients. This mismatch motivates explicit feature–coefficient alignment.

\section{Transform-Aligned Learned Features}
\label{sec:talf_formulation}

Our goal is to construct coefficient-specific representations for probability modeling. We establish two theoretical
properties in a general transform-block setting.  First, aligned features
exactly represent the first-order term of a shared nonlinear signal predictor
followed by the transform, with a second-order bound on the nonlinear
remainder.  Second, orthonormality preserves reconstruction error, enabling
coefficient-domain rate--distortion learning.  

\subsection{Transform-Block Formulation}
\label{sec:transform_coded_attribute_compression}

Consider a local transform block
$\mathcal B_\nu=\{1,\ldots,m_\nu\}$ with scalar signal samples
$\mathbf x_\nu=[x_{\nu,1},\ldots,x_{\nu,m_\nu}]^{\top}$.  The block may be
formed by any deterministic construction available at both codec ends.
We present one signal channel; multiple channels can be processed separately
or conditionally.  A known analysis transform produces
\begin{equation}
\mathbf c_\nu=\mathbf T_\nu\mathbf x_\nu,
\label{eq:talf_local_transform}
\end{equation}
where $\mathbf T_\nu\in\mathbb R^{m_\nu\times m_\nu}$ and
coefficient indices in $\mathbf c_\nu$ start at zero.

For the analysis below, we consider orthonormal transforms that separate a
constant signal component from its detail components.  These properties are standard in many classical transform-coding constructions~\citep{goyal2001transform,mallat2009wavelet}.  
With the low-pass, or DC, basis first and the remaining local detail, or AC,
rows denoted by $\mathbf T_{\nu,\mathrm{AC}}$, we write
\begin{equation}
\mathbf T_\nu^{\top}\mathbf T_\nu=\mathbf I,
\qquad
\mathbf T_{\nu,\mathrm{AC}}\mathbf1=\mathbf0.
\label{eq:talf_compatible_transform}
\end{equation}
The first row is $\mathbf d_\nu^{\top}=\mathbf1^{\top}/\sqrt{m_\nu}$,
so the low-pass coefficient equals $\sqrt{m_\nu}$ times the block mean.
A constant signal has zero detail coefficients.  The detail basis may be any
orthonormal complement of the constant direction.

We use unweighted signals here to expose the alignment principle.
Appendix~\ref{app:nonlinear_alignment_proof} extends the analysis to weighted
coordinates, and Section~\ref{sec:talf} specializes it to point-count-weighted
hierarchical attribute coding.

\subsection{Transform Alignment of Learned Features}
\label{sec:feature_coefficient_misalignment}

For each element $i\in\mathcal B_\nu$, an encoder extracts
$\mathbf h_{\nu,i}=\mathcal E_\theta(i,\mathcal C_\nu)\in\mathbb R^d$,
where $\mathcal C_\nu$ denotes any context available at both codec ends.
Stack these features in the same element order as the signal:
\begin{equation*}
\mathbf H_\nu
=[\mathbf h_{\nu,1},\ldots,\mathbf h_{\nu,m_\nu}]^{\top}
\in\mathbb R^{m_\nu\times d}.
\end{equation*}
Rows of $\mathbf H_\nu$ identify block elements, whereas entries of
$\mathbf{c}_\nu$ identify transform modes that generally combine multiple
elements.  Directly assigning element features to coefficient slots therefore
leaves a known basis change implicit, which we call
\emph{feature--coefficient misalignment}.

TALF removes this mismatch by applying the same analysis transform
to every feature channel:
\begin{equation}
\mathbf{Z}_\nu
=\mathbf{T}_\nu\mathbf{H}_\nu
\in\mathbb{R}^{m_\nu\times d}.
\label{eq:talf_alignment}
\end{equation}
The $k$-th row $\mathbf{z}_{\nu,k}^{\top}$ is paired with coefficient
$c_{\nu,k}$.  Each aligned feature provides an informative representation
for learned coefficient modeling, specifically coefficient prediction and
residual probability estimation.

\begin{figure}[t]
  \centering
  \includegraphics[width=0.95\textwidth]{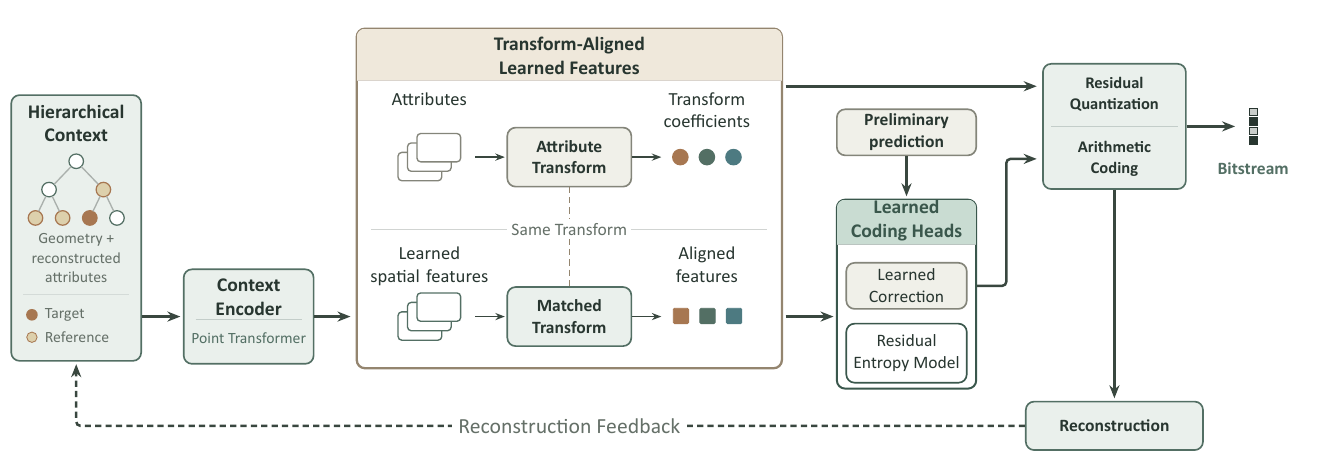}
  \caption{Overview of a TALF-based point-cloud attribute codec. Learned spatial features are aligned with attribute coefficients to guide coefficient prediction and residual entropy modeling. 
  }
  \label{fig:talf_overview}
\end{figure}

\subsection{Theoretical Analysis and Coefficient-Domain Learning}
\label{sec:why_transform_alignment_helps}
\label{sec:coefficient_domain_learning}

\paragraph{Local first-order alignment.}
We first establish why transform-aligned features provide a suitable
representation for coefficient prediction.
We fix a block and omit the subscript $\nu$ throughout this subsection.
Assume that its signal samples admit one common nonlinear function $f$:
\[
x_i=f(\mathbf h_i)+\epsilon_i,
\]
where $\epsilon_i$ is the residual not
explained by $f$.  The function may vary across blocks and
is assumed to be twice continuously differentiable in a neighborhood of the
convex hull of the block's features.  This regularity condition holds for
networks with smooth activations such as GELU~\citep{hendrycks2016gelu}.
Define the feature center as
$\bar{\mathbf h}=m^{-1}\sum_i\mathbf h_i$.
Expanding $f$ about $\bar{\mathbf h}$ and stacking the elements gives
\begin{equation*}
\begin{aligned}
\mathbf{x}
&=\mathbf H
\underbrace{\nabla f(\bar{\mathbf h})}_{\mathbf g}
+\underbrace{\left[
f(\bar{\mathbf h})
-\nabla f(\bar{\mathbf h})^{\top}\bar{\mathbf h}
\right]}_{\xi}\mathbf1
+\boldsymbol{\tau}
+\boldsymbol{\epsilon}\\
&=\mathbf H\mathbf g
+\xi\mathbf1
+\boldsymbol\tau
+\boldsymbol\epsilon,
\end{aligned}
\end{equation*}
Here $\boldsymbol{\tau}$ is the second-order Taylor remainder.  Since
$\mathbf{T}_{\mathrm{AC}}
\mathbf{1}=\mathbf{0}$, the shared intercept contributes
only to the low-pass coefficient.  Applying the detail rows to this expansion
therefore yields
\begin{equation}
\boxed{
\mathbf{c}_{\mathrm{AC}}
=\mathbf{Z}_{\mathrm{AC}}\mathbf{g}
+\mathbf{T}_{\mathrm{AC}}
(\boldsymbol{\tau}+\boldsymbol{\epsilon})
},
\label{eq:talf_nonlinear_alignment}
\end{equation}
where $\mathbf{Z}_{\mathrm{AC}}$ contains the detail rows of
$\mathbf{Z}$, and the remaining error comprises the transformed Taylor remainder and unexplained residual.
For a globally shared affine predictor $f(\mathbf h)=\mathbf g^{\top}\mathbf h+\xi$,
the Taylor remainder vanishes and $\mathbf g$ is identical across blocks.
A shared linear head on the aligned features then exactly reproduces the
transformed detail predictions.

For nonlinear $f$, if $\|\nabla^2f(\mathbf h)\|_2\leq\kappa$ along every segment
from $\bar{\mathbf h}$ to an element feature, the nonlinear discrepancy obeys
\begin{equation}
\left\|
\mathbf{T}_{\mathrm{AC}}
\boldsymbol{\tau}
\right\|_2
\leq
\frac{\kappa}{2}
\left(
\sum_i
\|\mathbf h_i-\bar{\mathbf h}\|_2^4
\right)^{1/2}.
\label{eq:talf_nonlinear_remainder_bound}
\end{equation}
The higher-order contribution is therefore small when the function has low
local curvature and the within-block features have limited dispersion.
Aligned features thus exactly represent the first-order prediction term,
with a bounded higher-order remainder, providing a structural basis for
learned coefficient modeling. In practice, a globally shared nonlinear head
approximates these block-dependent prediction relationships.
Appendix~\ref{app:nonlinear_alignment_proof}
provides the complete derivation.

\paragraph{Coefficient-domain rate--distortion objective.}
We next formulate the training objective in the coefficient domain.
Orthonormality preserves squared reconstruction error independently of the local prediction model. For any reconstruction $\hat{\mathbf x}$, with
$\hat{\mathbf c}=\mathbf T\hat{\mathbf x}$,
Parseval's identity gives
\begin{equation}
\|\mathbf x-\hat{\mathbf x}\|_2^2
=\|\mathbf c-\hat{\mathbf c}\|_2^2.
\label{eq:talf_parseval}
\end{equation}
Within each block, fixing the low-pass reconstruction makes detail-coefficient distortion equivalent to signal-domain distortion up to a constant term.
We therefore optimize the following coefficient-domain rate–distortion objective:
\begin{equation}
\mathcal{L}_{\mathrm{RD}}
=D_{\mathrm{coeff}}+\beta R
=\frac{1}{|\mathcal{K}|}
\sum_{k\in\mathcal{K}}
\left[
(c_k-\hat{c}_k)^2
+\beta\left(-\log_2 p(q_k\mid\mathcal{C}_k)\right)
\right],
\label{eq:talf_rd_objective}
\end{equation}
where $\mathcal{K}$ indexes the coefficients modeled by the learned
components, $q_k$ is the transmitted quantization index, $\mathcal{C}_k$ is
the coding context available at both codec ends, and $\beta>0$ controls the
rate--distortion trade-off.  This
coefficient-wise formulation associates every aligned feature with its target
coefficient, reconstruction error, and estimated rate.  The alignment
principle is independent of the context encoder, prediction head, entropy
model, and the choice of basis within the transform setting above.

\section{TALF-Based Attribute Codec}
\label{sec:talf}
\label{sec:talf_overview}

We instantiate the alignment principle of
Section~\ref{sec:talf_formulation} in a hierarchical
attribute codec. As shown in Figure~\ref{fig:talf_overview}, a context encoder extracts node features from hierarchical context, TALF aligns these features with the attribute coefficients, and learned coding heads predict coefficients and their residual distribution parameters. Reconstructed attributes then provide context
for subsequent coding steps. 

\subsection{Hierarchical Context and Aligned Feature Construction}
\label{sec:talf_context_transform}

We organize the point-cloud geometry using an octree, with the occupied children of each parent forming a transform block. Each child node \(u_i\) represents \(w_i\) points, and its attribute \(a_i\) is their average attribute.
With $\mathbf a_\nu=[a_1,\ldots,a_{m_\nu}]^{\top}$ and
$\mathbf M_\nu=\operatorname{diag}(w_1,\ldots,w_{m_\nu})$, the transform
input is $\mathbf x_\nu=\mathbf M_\nu^{1/2}\mathbf a_\nu$.
Each entry $x_i=\sqrt{w_i}\,a_i$ is the DC coefficient propagated from an
internal child's subtree, or the original attribute at a leaf where $w_i=1$.
This corresponds to $\mathbf S_\nu=\mathbf M_\nu^{1/2}$ in
Appendix~\ref{app:nonlinear_alignment_proof}.

For an occupied child node $u_i$ in block $\mathcal{B}_\nu$, we construct
a reference set
\(\mathcal{R}(u_i)=\mathcal{R}_{\mathrm{par}}(u_i)
\cup\mathcal{R}_{\mathrm{cur}}(u_i)\).
We first select nearby parent-level nodes and replace those whose child
attributes have already been reconstructed with their occupied children.
$\mathcal{R}_{\mathrm{par}}$ contains the remaining parent-level references,
and $\mathcal{R}_{\mathrm{cur}}$ contains the replacement child nodes.
Each reference contains relative coordinates, reconstructed attributes, and its own reconstructed prediction residual.
Higher-level coordinates are averages of child coordinates weighted by point counts.

Our implementation uses an adaptive Point Transformer
encoder~\citep{you2024efficient}. Query-masked attention combines the reference features with their relative
positions while suppressing padded entries. The resulting features are aggregated and mapped to a representation
$\mathbf{h}_{u_i}=\mathcal{E}_{\theta}(u_i,\mathcal{R}(u_i))$.
To match these transform inputs, we weight each learned
representation by $\sqrt{w_i}$ and stack the results in the same element order to form
$\mathbf{H}_\nu$.
This is the point-count-weighted
specialization of $\mathbf H_\nu$ in
Section~\ref{sec:feature_coefficient_misalignment}.

As a concrete instantiation of the general transform in
Section~\ref{sec:transform_coded_attribute_compression}, we use a
normalized graph-Laplacian basis~\citep{chen2020point} for both attribute
analysis and feature alignment. The graph construction and basis
conventions are specified in Appendix~\ref{app:transform_implementation}.
RAHT~\citep{de2016compression} provides an alternative basis
through the same interface.
Only features corresponding to valid AC coefficients enter the learned
heads; DC coefficients follow the hierarchical propagation rule.

\subsection{Coefficient Prediction and Residual Entropy Modeling}
\label{sec:talf_prediction_entropy}

Each aligned feature $\mathbf{z}_{\nu,k}$ supplies the context for two coefficient-specific quantities: a coefficient prediction and the parameters of its residual entropy model.

\paragraph{Explicit coefficient prediction.}
We use a deterministic preliminary predictor together with a learned
correction. For each occupied child node $u_i$, we interpolate the
reconstructed attributes of the three nearest reference nodes in
$\mathcal{R}(u_i)$ using inverse-distance weighting. Collecting these node-average
attribute predictions gives $\mathbf{a}_\nu^{\mathrm{pre}}$.
Square-root weighting followed by the attribute transform yields the
preliminary coefficient prediction
\[
\boldsymbol{\mu}_\nu^{\mathrm{pre}}
=
\mathbf{T}_\nu\mathbf{M}_\nu^{1/2}\mathbf{a}_\nu^{\mathrm{pre}}.
\]
The coefficient-prediction head $g_\mu$ produces a learned correction from
the aligned feature. Adding this correction to the preliminary prediction
gives the final coefficient prediction $\mu_{\nu,k}^{\mathrm{p}}$:
\begin{equation}
\mu_{\nu,k}^{\mathrm{p}}
=\mu_{\nu,k}^{\mathrm{pre}}+\mu_{\nu,k}^{\mathrm{corr}},
\qquad
\mu_{\nu,k}^{\mathrm{corr}}=g_{\mu}(\mathbf{z}_{\nu,k}).
\label{eq:talf_coefficient_prediction}
\end{equation}
For channel $\chi$ with quantization step $\Delta_\chi$, the quantized
residual index $q_{\nu,k}$ and reconstructed coefficient $\hat{c}_{\nu,k}$
are given by
\begin{equation}
q_{\nu,k}
=\mathcal{Q}\!\left(
\frac{c_{\nu,k}-\mu_{\nu,k}^{\mathrm{p}}}{\Delta_\chi}
\right),
\qquad
\hat{c}_{\nu,k}
=\mu_{\nu,k}^{\mathrm{p}}+\Delta_\chi\cdot q_{\nu,k}.
\label{eq:talf_residual_quantization}
\end{equation}
Here $\mathcal{Q}$ is the scalar quantizer in index units.
Unlike a location parameter used solely for entropy modeling, the explicit prediction affects both the transmitted residual index and the reconstructed coefficient.
Appendix~\ref{app:explicit_mu_analysis} further analyzes these two effects through the integer and fractional components of the learned correction.

\paragraph{Zero-centered residual entropy modeling.}
We model the residual indices with a zero-centered discretized Laplace
distribution. The entropy head predicts only a positive
scale parameter $b_{\nu,k}$, assigning probability and estimated rate
\[
p_{\nu,k}(q)
=\int_{q-\frac12}^{q+\frac12}
\operatorname{Lap}(t;0,b_{\nu,k})\,\mathrm{d}t,
\qquad
R_{\nu,k}=-\log_2 p_{\nu,k}(q_{\nu,k}).
\]
After transform alignment, a shared MLP maps each
$\mathbf{z}_{\nu,k}$ to a compact hidden representation, from which lightweight
heads predict $\mu_{\nu,k}^{\mathrm{corr}}$ and $b_{\nu,k}$. 
For color, corrections are predicted jointly, while scale prediction
is conditioned on previously decoded channel indices in the order
$\mathrm{Cb}\rightarrow\mathrm{Cr}\rightarrow\mathrm{Y}$.

The learned components are trained with the coefficient-domain
rate--distortion objective in \Eqref{eq:talf_rd_objective}.
The forward pass uses hard quantization indices, with a straight-through
estimator~\citep{bengio2013estimating} for backpropagation.
We also evaluate conventional run-length coding of the residual indices,
as reported in
Section~\ref{sec:main_rd_complexity}.

\subsection{Coding Procedure and Quantization-Step Control}
\label{sec:talf_qs_compatible}

Coding and reconstruction proceed from coarse to fine through the octree, starting from the root DC. Combining the AC coefficients reconstructed by
\Eqref{eq:talf_residual_quantization} with the DC supplied by the
coarser level gives $\hat{\mathbf{c}}_\nu$. The node-average attributes
$\hat{\mathbf{a}}_\nu
=\mathbf{M}_\nu^{-1/2}\mathbf{T}_\nu^{\top}\hat{\mathbf{c}}_\nu$
then provide context for subsequent blocks and finer levels. At the leaf level, they constitute the final point cloud reconstruction.

\paragraph{Conversion between quantization units.}
Converting learned coefficient corrections and entropy scales between quantization units enables a single model to operate across QS values with explicit quantization control.
Let $\Delta_\chi^{\mathrm r}$ and $\Delta_\chi^{\mathrm t}$ denote the
reference training QS and target QS, respectively. In implementation, the
prediction head parameterizes the correction by $\mu_\chi^{q,\mathrm r}$ in
units of the reference quantization step, so that
$\mu_\chi^{\mathrm{corr}}=\Delta_\chi^{\mathrm r}\cdot\mu_\chi^{q,\mathrm r}$.
With $b_\chi^{\mathrm r}$ denoting the predicted scale in the same units,
the target-QS parameters are
\begin{equation}
\rho_\chi=\frac{\Delta_\chi^{\mathrm r}}{\Delta_\chi^{\mathrm t}},
\qquad
\mu_\chi^{q,\mathrm t}=\rho_\chi\mu_\chi^{q,\mathrm r},
\qquad
b_\chi^{\mathrm t}=\rho_\chi b_\chi^{\mathrm r}.
\label{eq:talf_qs_conversion}
\end{equation}
This conversion preserves the coefficient-domain correction.
For cross-channel conditioning, the decoded symbol $q_\chi^{\mathrm t}$ is
rescaled to reference units before embedding as
$q_\chi^{\mathrm r}=q_\chi^{\mathrm t}/\rho_\chi$.

\section{Experiments}

We now present the experimental settings, compression performance, and ablation studies.

\subsection{Experimental Setup}
\label{sec:experimental_setup}

\paragraph{Datasets.}
We evaluate color compression on ScanNet~\citep{dai2017scannet}, using
1,513 scans for training and 100 scans for testing.
The geometry is quantized at a spatial resolution of $2$~mm, preserving point uniqueness and one-to-one correspondence with the associated colors. Reflectance experiments use the KITTI
LiDAR dataset~\citep{geiger2012we} and Ford~\citep{pandey2011ford}. KITTI sequences 00--10 are used
for training and 11--21 for testing; Ford sequence 01 is used for training,
with sequences 02 and 03 reserved for testing. Both reflectance datasets
are voxelized at a resolution of $1$~mm.

\paragraph{Baselines and metrics.}
Our default baseline, denoted G-PCCv33,
is TMC13v33~\citep{TMC13v33} configured with predictive RAHT.
We compare with 3DAC~\citep{fang2022_3dac}, Unicorn~\citep{wang2025unicorn},
and DeepRAHT~\citep{fu2026deepraht}. We retrain 3DAC on the same training
data as TALF and evaluate it on our test sets. We report the released Unicorn reflectance results with their original Unicorn-G-PCC (RAHT21) anchor for reference. For DeepRAHT, we
evaluate the executable released in its GitHub repository on our color
test set without retraining.

Attribute bitrate is reported in bits per input point (bpp). We report PSNR-Refl for reflectance and PSNR-YCbCr for color, with color MSE computed by weighting the Y, Cb, and Cr channel MSEs in a 6:1:1 ratio. Per-channel color PSNR is reported in the appendix. BD-Rate is computed using a
Bj{\o}ntegaard fit over the common quality interval~\citep{bjontegaard2001calculation}.

\paragraph{Implementation.}

Experiments use an NVIDIA Tesla T4 GPU and an Intel Xeon Gold 6248 CPU
at 2.50~GHz. For each dataset, a single model is evaluated at 13 base QS settings from 4 to 64 using the conversion in Section~\ref{sec:talf_qs_compatible}.
Network and training configurations are given in
Table~\ref{tab:implementation_config}, and the causal batching procedure
is described in Appendix~\ref{app:coding_evaluation}.

\subsection{Overall Rate--Distortion Performance}
\label{sec:main_rd_complexity}

\begin{table*}[!t]
\centering
\caption{
Comparisons of coding performance using average BD-Rate (\%) against the baseline methods and runtime (s/frame). G-PCCv33 is the BD-Rate anchor, except for Unicorn$^{\dagger}$, which uses the reported Unicorn-G-PCC (RAHT21) anchor.
}
\label{tab:average_bd_results}
\label{exp:main}
\noindent
\begin{minipage}[t]{0.46\textwidth}
\vspace{0pt}
\centering
\textbf{(a) Reflectance compression}
\par\vspace{2pt}

\footnotesize
\setlength{\tabcolsep}{4pt}
\begin{tabular}{lrrr}
\toprule
Dataset & 3DAC & Unicorn & TALF \\
\midrule

Ford
& $+1.81$
& $-5.63^{\dagger}$
& $\mathbf{-20.93}$ \\

KITTI
& $+9.89$
& $-4.68^{\dagger}$
& $\mathbf{-14.01}$ \\

\bottomrule
\end{tabular}
\par\vspace{2pt}
\end{minipage}%
\hfill%
\begin{minipage}[t]{0.53\textwidth}
\vspace{0pt}
\centering
\textbf{(b) Color compression}
\par\vspace{2pt}

\footnotesize
\setlength{\tabcolsep}{1.0pt}
\begin{tabular}{lccccc}
\toprule
Dataset & 3DAC & DeepRAHT &
\shortstack{TALF-\\Neural} & \shortstack{TALF-\\RL} &
\shortstack{TALF-\\Hybrid} \\
\midrule

ScanNet
& $+3.85$
& $-6.65$
& $-9.71$
& $-12.26$
& $\mathbf{-15.07}$
\\

\bottomrule
\end{tabular}
\end{minipage}

\vspace{8pt}

\textbf{(c) Runtime (Enc./Dec.)}
\vspace{2pt}

\footnotesize
\setlength{\tabcolsep}{7pt}
\begin{tabular}{lccc}
\toprule
Method & Ford & KITTI & ScanNet \\
\midrule
G-PCCv33 & 1.5 / 0.8 & 0.6 / 0.3 & 1.8 / 1.4 \\
3DAC  & 3.8 / 4.5 & 3.6 / 4.9 & 4.3 / 5.0 \\
DeepRAHT & -- & -- & 3.1 / 3.1 \\
TALF  & 7.9 / 7.7
      & 9.0 / 8.9
      & 8.5 / 8.4 \\
\bottomrule
\end{tabular}
\end{table*}

\begin{figure}[!t]
    \centering

    \begin{subfigure}[t]{0.333\textwidth}
        \centering
        \includegraphics[viewport=0 0 181.5 139.89,clip,width=\linewidth]{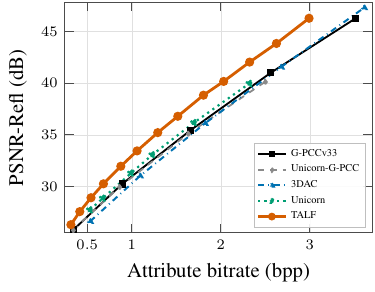}
        \captionsetup{margin={20pt,0pt}}
        \caption{Ford.}
        \label{fig:rd_ford}
    \end{subfigure}%
    \hfill%
    \begin{subfigure}[t]{0.333\textwidth}
        \centering
        \includegraphics[viewport=0 0 181.5 139.89,clip,width=\linewidth]{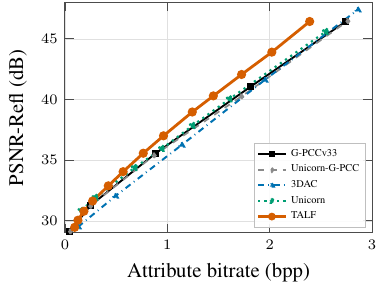}
        \captionsetup{margin={20pt,0pt}}
        \caption{KITTI.}
        \label{fig:rd_kitti}
    \end{subfigure}%
    \hfill%
    \begin{subfigure}[t]{0.333\textwidth}
        \centering
        \includegraphics[viewport=0 0 181.5 139.89,clip,width=\linewidth]{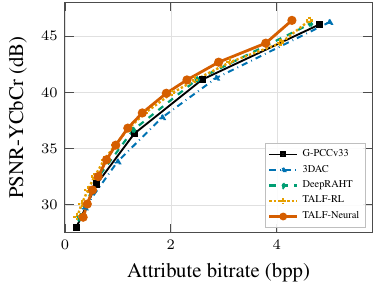}
        \captionsetup{margin={24pt,0pt}}
        \caption{ScanNet.}
        \label{fig:rd_scannet}
    \end{subfigure}

    \caption{
        Rate--distortion curves for Ford and KITTI reflectance and ScanNet
        color.
    }
    \label{fig:overall_rd_results}
\end{figure}

Figure~\ref{fig:overall_rd_results} and
Table~\ref{tab:average_bd_results} show consistent rate--distortion
improvements over G-PCCv33 across three datasets, with bitrate
savings reaching $20.93\%$ on Ford. The gains extend from outdoor LiDAR
reflectance to indoor color scans, demonstrating the effectiveness of the
codec across different attribute types and scene conditions.

On ScanNet, we distinguish three variants with identical coefficient
predictions and reconstructions: TALF-Neural uses the learned residual
entropy model, TALF-RL uses conventional run-length coding, and TALF-Hybrid uses learned entropy coding for $\mathrm{QS}\leq16$ and
run-length coding otherwise, using a validation-selected threshold fixed for
all test samples.
All three variants outperform 3DAC and DeepRAHT in average BD-Rate, with
TALF-Hybrid achieving the largest savings. As shown in
Table~\ref{tab:average_bd_results}(c), G-PCCv33 is the fastest method,
while TALF has moderately higher runtimes than 3DAC and DeepRAHT.

\subsection{Analysis of Transform Alignment}
\label{sec:alignment_ablation}
\label{sec:transform_basis_evaluation}

We examine the first-order approximation, the effectiveness of explicit
feature alignment, and its applicability across transform bases on Ford.

\begin{table*}[t]
\centering
\caption{Analysis of transform alignment on Ford.
(a) Residual energy reduction is relative to zero-innovation prediction.
(b) Prediction MSE is evaluated on valid AC coefficients before quantization.
(c) BD-Rate is relative to G-PCCv33.
All reductions, increases, and BD-Rate values are in \%.}
\label{tab:ford_theoretical_validation}

\textbf{(a) Assessing the first-order approximation}
\vspace{2pt}

{\footnotesize
\setlength{\tabcolsep}{5pt}
\begin{tabular}{lcccc}
\toprule
Predictor
&
Nonlinearity
&
Parameters
&
Innovation MSE
&
\shortstack{Residual energy\\reduction}
\\
\midrule

Shared linear predictor
& None
& 513
& 139.2518
& 30.37 \\

Shared one-hidden-layer MLP
& GELU
& 263,169
& 135.2843
& 32.35 \\

\bottomrule
\end{tabular}
}

\vspace{8pt}

\noindent
\begin{minipage}[t]{0.53\textwidth}
\vspace{0pt}
\centering
\textbf{(b) Feature-mapping ablation relative to TALF}
\par\vspace{2pt}

\footnotesize
\setlength{\tabcolsep}{4pt}
\begin{tabular}{lcc}
\toprule
Feature mapping
& \shortstack{Prediction MSE\\increase}
& \shortstack{BD-Rate vs.\\TALF} \\
\midrule

Weighted centering
& $+30.45$
& $+13.32$ \\

Learned transform
& $+28.11$
& $+10.17$ \\

\bottomrule
\end{tabular}
\end{minipage}%
\hfill%
\begin{minipage}[t]{0.45\textwidth}
\vspace{0pt}
\centering
\textbf{(c) Transform-basis comparison}
\par\vspace{2pt}

\footnotesize
\setlength{\tabcolsep}{2.5pt}
\begin{tabular}{lcc}
\toprule
Method & Transform basis & BD-Rate \\
\midrule
TALF-RAHT
& RAHT
& $-20.35$ \\
TALF
& Graph
& $\mathbf{-20.93}$ \\
\bottomrule
\end{tabular}
\end{minipage}

\end{table*}

\paragraph{Assessing the first-order approximation.}
We examine whether a single linear predictor shared across child nodes
captures most of the prediction benefit provided by a nonlinear model.
Specifically, we freeze the feature encoder and compare a shared
linear predictor with a nonlinear MLP predictor for predicting the child-attribute
innovations after the deterministic preliminary predictor.
As shown in Table~\ref{tab:ford_theoretical_validation}(a), after convergence, the shared
linear predictor yields approximately $94\%$ of the MLP's residual
energy reduction with approximately $500\times$ fewer trainable parameters.
These results suggest that linear prediction accounts for most of the
observed gain, supporting the first-order analysis in
Section~\ref{sec:why_transform_alignment_helps}.

\paragraph{Feature-mapping ablation.}
We replace TALF's explicit feature transform with two alternative
mappings while keeping the remaining coding framework unchanged.
The \emph{weighted-centering} variant forms child representations as
$\widetilde{\mathbf h}_{u_i}=\sqrt{w_i}(\mathbf h_{u_i}-\bar{\mathbf h}_w)$,
where $\bar{\mathbf h}_w=\sum_j w_j\mathbf h_{u_j}/\sum_j w_j$.
The point-count weights $w_i$ are defined in
Section~\ref{sec:talf_context_transform}.
The first $m_\nu-1$ centered features are paired with AC coefficients in
coding order without applying $\mathbf T_\nu$.
The \emph{learned-transform} variant instead uses a shared
nonlinear mapping across child features, allowing data-driven
alignment without imposing the codec's transform basis.

Table~\ref{tab:ford_theoretical_validation}(b) shows that weighted
centering and the learned mapping both increase prediction MSE, with
BD-Rate penalties of $13.32\%$ and $10.17\%$, respectively, relative to TALF.
These results indicate that removing the constant component alone is insufficient, while learning the mapping does not fully recover the benefit of explicit transform alignment.

\paragraph{Applicability across transform bases.}
To examine dependence on the transform basis, we instantiate the same
alignment interface with RAHT.
The corresponding RAHT
operator~\citep{taquet2020dyadicraht} is applied to both attribute signals and learned features,
while the other coding components remain unchanged.
As shown in Table~\ref{tab:ford_theoretical_validation}(c), TALF-RAHT
achieves $20.35\%$ BD-Rate savings over G-PCCv33 on Ford, only
$0.58$ percentage points below TALF.
These results support the applicability of the alignment interface
across the two tested transform bases.

\subsection{Explicit Coefficient Correction and Entropy Coding}
\label{sec:prediction_backend_analysis}

\paragraph{Effect of explicit coefficient correction.}
We decompose the learned correction in quantization-index units as
$\mu^q=\mu_n^q+\mu_f^q$, where $\mu_n^q=\operatorname{round}(\mu^q)$ and
$\mu_f^q$ is the remaining fractional component. The No $\mu$ variant uses the
deterministic preliminary predictor without learned coefficient correction;
Integer $\mu_n^q$ adds the grid-aligned
learned correction; Full $\mu$ also includes its fractional part.
Each variant uses the same trained model and codec configuration but advances its own
reconstructed causal context and produces an actual arithmetic-coded
bitstream across the evaluated QS range.

\begin{table}[t]
\centering
\caption{Ablation of explicit coefficient correction on Ford.
BD-Rate (\%) is reported relative to G-PCCv33,
No $\mu$, and Integer $\mu_n^q$.}
\label{tab:ford_mu_closed_loop_bd_rate}
\small
\setlength{\tabcolsep}{5pt}
\begin{tabular}{lccc}
\toprule
Prediction variant
& \shortstack{BD-Rate vs.\\G-PCCv33}
& \shortstack{BD-Rate vs.\\No $\mu$}
& \shortstack{BD-Rate vs.\\Integer $\mu_n^q$} \\
\midrule
No $\mu$   & -7.68  & 0.00   & -- \\
Integer $\mu_n^q$ & -15.50 & -8.46  & 0.00 \\
Full $\mu$  & -20.93 & -14.02 & -6.00 \\
\bottomrule
\end{tabular}
\end{table}

Without learned coefficient correction, No $\mu$ still models residual
uncertainty through conditional scales and achieves $7.68\%$ BD-Rate
savings over G-PCCv33 (Table~\ref{tab:ford_mu_closed_loop_bd_rate}).
Integer correction further recenters
the coded symbols. The fractional component shifts the reconstruction grid within a quantization step and enables learned adjustments to quantization decisions toward lower rate–distortion cost. Together, these effects yield an additional \(6.00\%\) BD-Rate reduction over Integer \(\mu_n^q\).
Appendix~\ref{app:explicit_mu_analysis} gives the corresponding local
symbol statistics and coefficient-error analysis.

\paragraph{Effect of the entropy-coding method.}
To examine entropy-backend dependence, we retain TALF's coefficient
prediction and feature alignment while switching between neural and
run-length coding~\citep{mpeg2020runlength}, keeping reconstructed coefficients fixed
(Figure~\ref{fig:overall_rd_results}(c)).
TALF-Neural uses fewer bits at fine QS, whereas TALF-RL is more
efficient at coarse QS, where residuals are sparse.
Part of this gap arises from the finite-precision entropy coder.
Our 16-bit arithmetic CDF uses a
2049-symbol alphabet, capping the probability of any symbol at $31/32$.
Consequently, even all-zero color residuals incur an approximate rate floor of \(0.137\) bpp. For ScanNet scene 0707\_00 at the coarsest tested QS, the neural and run-length payloads are \(0.237\) and \(0.112\) bpp, respectively. The latter falls below this floor, showing that improved probability estimates alone cannot close the gap.

\section{Conclusion}
We presented Transform-Aligned Learned Features (TALF), which applies the attribute transform to align learned spatial representations with their coefficient targets. Our framework combines TALF, explicit coefficient prediction, and conditional residual entropy modeling under a unified coefficient-domain rate–distortion objective, while preserving quantization-step control. Results on reflectance and color show coding gains over conventional and learned baselines, with both the normalized graph basis and RAHT supporting the effectiveness of the alignment principle.

More broadly, this work suggests that attribute transforms can structure learned representations by making the known basis change explicit, while neural models learn contextual dependencies. The TALF alignment interface has the potential to accommodate different context encoders, transform bases, and entropy models. In future, we can exploit
this flexibility to develop probability models better suited to sparse
residuals and investigate alignment in other transform-based compression
settings, such as graph signals and 3D Gaussian Splatting (3DGS) attributes.

\subsection*{AI Use Statement}
We used generative AI tools to assist with language editing, literature organization, figure and table preparation, and code implementation and debugging. The authors take responsibility for the manuscript, code, and reported results.
\subsection*{Reproducibility Statement}
To support reproducibility, we describe the model architecture, coding
procedure, and training settings in the main text. Our experiments use
publicly available datasets, with data splits, preprocessing, and evaluation
protocols specified in Section~\ref{sec:experimental_setup}. We state the
assumptions underlying our theoretical analysis and provide the corresponding
derivations. The appendix supplements these descriptions with transform
implementation details, analysis of the learned coefficient correction,
and dataset-average rate--distortion results.
We will publicly release our code, trained model weights, and data-preparation
scripts to enable independent verification and further research.

\bibliography{iclr2027_conference}
\bibliographystyle{iclr2027_conference}

\clearpage
\appendix
\raggedbottom
\section{Appendix}

This appendix provides implementation details for the TALF-based codec,
including transform construction, network architecture, training, and evaluation.

\subsection{Transform Construction}
\label{app:transform_implementation}
\label{app:transform_construction}

Our main instantiation uses a normalized graph-Laplacian basis~\citep{chen2020point,pavez2020ragft}.
For a block $\mathcal{B}_\nu$, its occupied children form a local graph
with edge set $\mathcal{E}_\nu$. With child positions $\mathbf{r}_i$,
the Gaussian adjacency weights are
\[
\mathbf{A}_\nu[i,j]
=
\begin{cases}
\displaystyle
\exp\!\left(
-\frac{\|\mathbf{r}_i-\mathbf{r}_j\|_2^2}{2\sigma_\nu^2}
\right), & (i,j)\in\mathcal{E}_\nu,\\[6pt]
0, & \text{otherwise},
\end{cases}
\]
where $\sigma_\nu$ is proportional to the cell size at the current level.
Child-node positions are computed as point-count-weighted averages of their
constituent point coordinates. We connect two occupied octants when their cells share a
face. We set \(\sigma_\nu\) equal to the current child-cell size.
Let $w_i$ be the number of source points represented by child $u_i$,
and define the diagonal node-weight matrix
$\mathbf{M}_\nu=\operatorname{diag}(w_1,\ldots,w_{m_\nu})$.
With the degree matrix
$\mathbf{D}_\nu=\operatorname{diag}(\mathbf{A}_\nu\mathbf{1})$, we construct
\[
\widetilde{\mathbf{L}}_\nu
=\mathbf{M}_\nu^{-1/2}
(\mathbf{D}_\nu-\mathbf{A}_\nu)\mathbf{M}_\nu^{-1/2}
=\mathbf{U}_\nu\boldsymbol{\Lambda}_\nu\mathbf{U}_\nu^{\top},
\qquad
\mathbf{T}_\nu=\mathbf{U}_\nu^{\top}.
\]
The eigenvectors are orthonormal, so
$\mathbf{T}_\nu^{\top}\mathbf{T}_\nu=\mathbf{I}$.
The first row produces the low-pass coefficient and is fixed to
\[
\mathbf{t}_{\nu,0}
=(\textstyle\sum_i w_i)^{-1/2}
[\sqrt{w_1},\ldots,\sqrt{w_{m_\nu}}],
\]
while the remaining rows produce local detail coefficients and follow
a deterministic sign and ordering convention.

Face connectivity does not necessarily produce a connected graph for the
occupied children. When the graph has multiple connected components, its
Laplacian has a multidimensional zero eigenspace. We fix the weighted
constant direction as the low-pass basis and construct the remaining
zero-eigenvalue vectors in its orthogonal complement; these vectors encode
differences between component averages. If no edges exist, the Laplacian
is zero, and the detail basis is obtained by orthogonal completion of the
prescribed low-pass vector. The transform remains orthonormal and invertible.
For repeated eigenvalues, we retain the deterministic eigenvector order
produced by the Jacobi solver, followed by fixed reorthogonalization and sign
canonicalization. The encoder and decoder derive the transform from the same
decoded geometry using identical numerical conventions, requiring no
additional basis information.

For node-average attributes $\mathbf{a}_\nu$ and raw encoder features
$\mathbf{F}_\nu$, the point-count-weighted transform inputs used by the codec are
$\mathbf{x}_\nu=\mathbf{M}_\nu^{1/2}\mathbf{a}_\nu$ and
$\mathbf{H}_\nu=\mathbf{M}_\nu^{1/2}\mathbf{F}_\nu$.
Applying the orthonormal transform gives
\[
\mathbf{c}_\nu=\mathbf{T}_\nu\mathbf{x}_\nu,
\qquad
\mathbf{Z}_\nu=\mathbf{T}_\nu\mathbf{H}_\nu.
\]
Reconstructed node-average attributes are recovered as
$\hat{\mathbf{a}}_\nu
=\mathbf{M}_\nu^{-1/2}\mathbf{T}_\nu^{\top}\hat{\mathbf{c}}_\nu$.

\subsection{First-Order Alignment and Reconstruction Error Preservation}
\label{app:nonlinear_alignment_proof}

We extend the two properties in
Section~\ref{sec:why_transform_alignment_helps} to weighted transform inputs:
exact alignment of the first-order prediction term with a bounded
higher-order remainder, and preservation of squared reconstruction error.
The unweighted formulation in Section~\ref{sec:talf_formulation} follows
by setting the weighting matrix to the identity.

\paragraph{Weighted transform formulation.}
Let $i=1,\ldots,m_\nu$ index the elements of block $\nu$.  Associate each
element with a positive weight $\alpha_{\nu,i}$ and collect
them in the diagonal matrix
\begin{equation*}
\mathbf{S}_\nu=\operatorname{diag}(\alpha_{\nu,1},\ldots,
\alpha_{\nu,m_\nu}).
\end{equation*}
For signal samples $\mathbf a_\nu$, define the weighted transform input
$\mathbf x_\nu=\mathbf S_\nu\mathbf a_\nu$ and coefficients
$\mathbf c_\nu=\mathbf T_\nu\mathbf x_\nu$.
In the point-count-weighted construction of Section~\ref{sec:talf},
$\mathbf S_\nu=\mathbf M_\nu^{1/2}$ and $\mathbf a_\nu$ contains
node-average attributes.  Weighting is applied to the input; the transform
$\mathbf T_\nu$ remains orthonormal.
Define
\begin{equation*}
\boldsymbol{\alpha}_\nu=\mathbf{S}_\nu\mathbf{1},
\qquad
\gamma_\nu=\|\boldsymbol{\alpha}_\nu\|_2.
\end{equation*}
Partition the analysis transform as
\begin{equation*}
\mathbf{T}_\nu=
\begin{bmatrix}
\mathbf{d}_\nu^{\top}\\
\mathbf{T}_{\nu,\mathrm{AC}}
\end{bmatrix},
\qquad
\mathbf{T}_\nu^{\top}\mathbf{T}_\nu=\mathbf{I}.
\end{equation*}
The first row is the low-pass basis, and the remaining rows form the detail
basis.  To separate a constant signal from its detail components after
weighting, the condition $\mathbf T_{\nu,\mathrm{AC}}\mathbf1=\mathbf0$
in \Eqref{eq:talf_compatible_transform} becomes
\begin{equation}
\mathbf T_{\nu,\mathrm{AC}}\boldsymbol\alpha_\nu=\mathbf0.
\label{eq:app_ac_annihilates_constant}
\end{equation}
Since an orthonormal
$\mathbf T_\nu$ has $m_\nu-1$ independent detail rows, the nullspace of
$\mathbf T_{\nu,\mathrm{AC}}$ is one-dimensional and is spanned by
$\mathbf d_\nu$.  Because $\boldsymbol\alpha_\nu\neq\mathbf0$,
\Eqref{eq:app_ac_annihilates_constant} holds if and only if
$\mathbf d_\nu=\pm\boldsymbol\alpha_\nu/\gamma_\nu$.  Fixing the positive
low-pass convention gives
\begin{equation}
\mathbf d_\nu=\frac{\boldsymbol\alpha_\nu}{\gamma_\nu}
=\frac{\mathbf S_\nu\mathbf1}{\|\mathbf S_\nu\mathbf1\|_2}.
\label{eq:app_constant_direction}
\end{equation}
Thus the low-pass basis follows the normalized weighted constant direction.

For a constant signal
$\mathbf a_\nu=a\mathbf1$, its weighted transform input need not
be elementwise constant.  Instead,
\begin{equation*}
\mathbf{x}_\nu
=\mathbf{S}_\nu\mathbf{a}_\nu
=a\boldsymbol{\alpha}_\nu
=a\gamma_\nu\mathbf{d}_\nu.
\end{equation*}
It lies exactly in the low-pass direction and therefore produces
\begin{equation*}
\mathbf T_\nu\mathbf S_\nu(a\mathbf1)
=
\begin{bmatrix}
a\gamma_\nu\\
\mathbf0
\end{bmatrix}.
\end{equation*}

\paragraph{Local first-order alignment.}
Stack the element features and define their weighted and aligned forms:
\begin{equation*}
\begin{aligned}
\mathbf F_\nu
&=[\mathbf h_{\nu,1},\ldots,\mathbf h_{\nu,m_\nu}]^{\top},\\
\mathbf H_\nu&=\mathbf S_\nu\mathbf F_\nu,
\qquad
\mathbf Z_\nu=\mathbf T_\nu\mathbf H_\nu.
\end{aligned}
\end{equation*}
Assume that the signal samples admit one common nonlinear function $f_\nu$:
\begin{equation*}
a_{\nu,i}=f_\nu(\mathbf h_{\nu,i})+\epsilon_{\nu,i}.
\end{equation*}
Here $\epsilon_{\nu,i}$ is the residual not explained by $f_\nu$.
The function may vary across blocks and is twice continuously
differentiable in a neighborhood of the convex hull of the block's features,
as in Section~\ref{sec:why_transform_alignment_helps}.
Use the weighted feature center
\begin{equation*}
\bar{\mathbf h}_\nu
=\frac{1}{\gamma_\nu^2}
\sum_{i=1}^{m_\nu}\alpha_{\nu,i}^2\mathbf h_{\nu,i},
\qquad
\mathbf g_\nu=\nabla f_\nu(\bar{\mathbf h}_\nu).
\end{equation*}
Taylor's theorem gives, for each element,
\begin{equation*}
f_\nu(\mathbf h_{\nu,i})
=f_\nu(\bar{\mathbf h}_\nu)
+\mathbf g_\nu^{\top}
(\mathbf h_{\nu,i}-\bar{\mathbf h}_\nu)
+\tau_{\nu,i}.
\end{equation*}
With
$\xi_\nu=f_\nu(\bar{\mathbf h}_\nu)
-\mathbf g_\nu^{\top}\bar{\mathbf h}_\nu$, stacking the elements yields
\begin{equation*}
\mathbf a_\nu
=\mathbf F_\nu\mathbf g_\nu
+\xi_\nu\mathbf1
+\boldsymbol\tau_\nu
+\boldsymbol\epsilon_\nu.
\end{equation*}
Multiplication by $\mathbf S_\nu$ sends the common intercept to
$\xi_\nu\boldsymbol\alpha_\nu$, which is removed exactly by every detail row
through \Eqref{eq:app_ac_annihilates_constant}.  Consequently,
\begin{equation}
\mathbf c_{\nu,\mathrm{AC}}
=\mathbf Z_{\nu,\mathrm{AC}}\mathbf g_\nu
+\mathbf T_{\nu,\mathrm{AC}}\mathbf S_\nu
(\boldsymbol\tau_\nu+\boldsymbol\epsilon_\nu).
\label{eq:app_exact_first_order_alignment}
\end{equation}
The aligned features therefore represent the first-order prediction term
exactly in the detail basis.  The remaining terms are the transformed Taylor
remainder and unexplained residual.  If
$\|\nabla^2f_\nu(\mathbf h)\|_2\leq\kappa_\nu$ on every line segment from
$\bar{\mathbf h}_\nu$ to an element feature, then
\begin{equation*}
|\tau_{\nu,i}|
\leq\frac{\kappa_\nu}{2}
\|\mathbf h_{\nu,i}-\bar{\mathbf h}_\nu\|_2^2.
\end{equation*}
Combining this bound with
$\|\mathbf T_{\nu,\mathrm{AC}}\mathbf v\|_2\leq\|\mathbf v\|_2$
for any vector $\mathbf v$ gives
\begin{equation}
\left\|
\mathbf T_{\nu,\mathrm{AC}}\mathbf S_\nu
\boldsymbol\tau_\nu
\right\|_2
\leq
\frac{\kappa_\nu}{2}
\left(
\sum_{i=1}^{m_\nu}\alpha_{\nu,i}^2
\|\mathbf h_{\nu,i}-\bar{\mathbf h}_\nu\|_2^4
\right)^{1/2}.
\label{eq:app_vector_remainder_bound}
\end{equation}
This bound controls the higher-order contribution, separately from
$\boldsymbol\epsilon_\nu$.  For fixed weights, this contribution is small
when the function has low local curvature and the block's features have
limited dispersion.  Setting $\mathbf S_\nu=\mathbf I$ recovers the feature
center and remainder bound in Section~\ref{sec:why_transform_alignment_helps}.

\paragraph{Reconstruction error preservation.}
Independently of the local prediction model, orthonormality preserves squared
reconstruction error.  For any signal reconstruction $\hat{\mathbf a}_\nu$,
let $\hat{\mathbf x}_\nu=\mathbf S_\nu\hat{\mathbf a}_\nu$ and
$\hat{\mathbf c}_\nu=\mathbf T_\nu\hat{\mathbf x}_\nu$.  Then
\begin{equation}
\|\mathbf S_\nu(\mathbf a_\nu-\hat{\mathbf a}_\nu)\|_2^2
=\|\mathbf x_\nu-\hat{\mathbf x}_\nu\|_2^2
=\|\mathbf c_\nu-\hat{\mathbf c}_\nu\|_2^2.
\label{eq:app_local_parseval}
\end{equation}
Thus coefficient-domain squared error equals weighted signal-domain squared
error, reducing to ordinary squared error when $\mathbf S_\nu=\mathbf I$.
For point-count weights, this is
$\sum_i w_i(a_{\nu,i}-\hat a_{\nu,i})^2$, the point-count-weighted
error of the node-average attributes.  Its aggregation across the hierarchy
is determined by the codec.  This identity supports the coefficient-domain
distortion term in Section~\ref{sec:coefficient_domain_learning} without
requiring the local nonlinear approximation.

\subsection{Network Architecture and Training}
\label{app:network_training}

Each reference is represented by relative coordinates, reconstructed
attributes, and its reconstructed prediction residual. For every target, the
relative coordinates of its first 32 references are divided by their maximum
absolute coordinate component. Reconstructed attributes
are centered by the preliminary prediction, and both the centered attributes
and reconstructed residuals are divided by 255. This gives five input channels
for reflectance and nine for color. Padded references are masked during
attention. The input projection maps these features to width 128. Five masked
attention blocks are followed by masked sum pooling and a
$128\!\rightarrow\!512$ adapter. The resulting child features are weighted and
transformed following Section~\ref{sec:talf_context_transform}.

A shared $512\!\rightarrow\!256$ MLP processes each transform-aligned feature,
followed by lightweight heads that predict the coefficient correction and
residual scale. Only features associated with local detail coefficients are
passed to these heads. The reflectance heads are linear maps from 256 to one.
For color, one linear head jointly predicts the three channel corrections. The
scale heads operate causally: the Cb scale head is evaluated first, followed by
the Cr scale head and then the Y scale head. For each decoded residual index,
we preserve its sign, apply $\log(1+|q|)$ to its
magnitude, and map the resulting scalar to a 32-dimensional embedding using a
linear layer followed by ReLU. Consequently, the Cb, Cr, and Y scale heads
receive 256, 288, and 320 inputs, respectively; each uses a 256-unit hidden
layer and a scalar softplus output.

We train a separate model for each dataset at a fixed reference QS using the
coefficient-domain rate--distortion objective. The context encoder, explicit
coefficient predictor, and residual entropy model are optimized jointly. At
most two million eligible blocks are sampled per epoch. A held-out subset of
each training split is used for validation, hyperparameter selection, and
checkpoint selection; the test sets are used only for final evaluation. The
rate weights, $\beta=1.5$ for reflectance and $\beta=18.0$ for color, are
selected empirically on the corresponding validation sets and then fixed
across all experiments. The learned path is enabled after skipping the three
coarsest octree levels, provided that the target has more than 16 valid
references; all other coefficients use the conventional path. During
evaluation, the same model is used across QS values through the conversion
described in Section~\ref{sec:talf_qs_compatible}.

\begin{table}[t]
\centering
\caption{Network and training configuration used for the reported models.}
\label{tab:implementation_config}
\footnotesize
\renewcommand{\arraystretch}{1.16}
\setlength{\tabcolsep}{3pt}
\begin{tabular}{p{0.29\linewidth}p{0.28\linewidth}p{0.34\linewidth}}
\toprule
Configuration & Reflectance & Color \\
\midrule
Maximum references per node & 32 & 32 \\
Attention blocks & 5 & 5 \\
Attention width & 128 & 128 \\
Output feature dimension & 512 & 512 \\
Input feature dimension & 5 & 9 \\
Shared MLP dimensions & $512\!\rightarrow\!256$ & $512\!\rightarrow\!256$ \\
Prediction head & $256\!\rightarrow\!1$ & $256\!\rightarrow\!3$ \\
Scale head & $256\!\rightarrow\!1$ & $256/288/320\!\rightarrow\!256\!\rightarrow\!1$ \\
Activations & ReLU; softplus scale & ReLU; softplus scale \\
\midrule
Optimizer & Adam & Adam \\
Initial learning rate & $10^{-4}$ & $10^{-4}$ \\
Learning-rate schedule & Constant & Constant \\
Training epochs & 40 & 40 \\
Training batch size & 128 (Ford), 128 (KITTI) & 128 \\
Reference QS & 4 & 4 \\
Rate weight $\beta$ & 1.5 & 18.0 \\
Max. blocks per epoch & $2\times10^6$ & $2\times10^6$ \\
Checkpoint selection & Minimum validation objective & Minimum validation objective \\
\bottomrule
\end{tabular}
\end{table}

\subsection{Coding and Evaluation Details}
\label{app:coding_evaluation}

The forward pass uses hard residual quantization, and coefficient
reconstruction follows \Eqref{eq:talf_residual_quantization}.
The quantizer uses a deadzone offset of $1/3$. The rate branch uses
$q_{\mathrm{rate}}=r/\Delta+\operatorname{sg}(q-r/\Delta)$, where
$r=c-\mu^{\mathrm{p}}$, $q$ is the hard index, and $\operatorname{sg}$ denotes stop
gradient. Thus, the forward value is the hard index while its backward
derivative with respect to the continuous residual index is one. In the
distortion branch, the dequantized hard residual is detached and added to the
predicted coefficient; gradients therefore pass through the explicit
prediction while the selected hard symbol remains fixed.

During encoding and decoding, coding proceeds from coarse to fine over the
octree levels. The root low-pass coefficient and all blocks at levels containing at most 128 transform blocks are processed sequentially
using the conventional run-length arithmetic backend. For each remaining
level, the blocks are divided in coding order into 256 contiguous sections
for reflectance and 64 for color. Processing is interleaved across sections: batching round $t$ collects the
$t$-th block from every nonempty section, so up to 256 reflectance blocks or
64 color blocks are evaluated in parallel.
After each coding step,
the reconstructed attributes are committed before the reference sets for the
next step are formed. The encoder and decoder replay the same level-wise
coding schedule to maintain identical causal contexts.


Evaluation uses the data splits and preprocessing described in
Section~\ref{sec:experimental_setup}. Dataset-level bitrate and MSE are
averaged across test sequences, or scans for ScanNet, and PSNR is computed
from the averaged MSE. Color MSE combines the Y, Cb, and Cr channel MSEs in
a 6:1:1 ratio. BD-Rate is calculated over the common quality interval using
the complete operating-point sets. Runtime is measured at QS~16 for the
learned-entropy variant. 

\subsection{Analysis of Explicit Coefficient Correction}
\label{app:explicit_mu_analysis}

We analyze how the integer and fractional components of the learned
coefficient correction affect residual coding and reconstruction.
We decompose the correction in quantization-index units as
$\mu^q=\mu_n^q+\mu_f^q$, where
$\mu_n^q=\operatorname{round}(\mu^q)$ is aligned with the integer quantization grid
and $\mu_f^q=\mu^q-\mu_n^q$ is the remaining fractional component.
The corresponding coefficient-domain components are
$\mu_n=\Delta\cdot\mu_n^q$ and $\mu_f=\Delta\cdot\mu_f^q$, giving
$\mu^{\mathrm{corr}}=\Delta\cdot\mu^q=\mu_n+\mu_f$.

\begin{table}[!htbp]
\centering
\caption{
Mechanistic analysis of the explicit coefficient correction on Ford. The integer-only and full corrections are compared with the no-correction path using the codec's actual quantizer.
Probabilities, zero-symbol ratios, and relative changes are reported in \%.
}
\label{tab:ford_mu_symbol_statistics}

\small
\renewcommand{\arraystretch}{1.08}
\setlength{\tabcolsep}{6pt}
\textbf{(a) Learned correction decomposition}\par\vspace{2pt}
\begin{tabular*}{295.1pt}{@{\extracolsep{\fill}}cccc@{}}
\toprule
QS & $\Pr(\mu_n^q\neq0)$ & $\mathbb{E}[|\mu_n^q|]$
& $\Pr(q_{\mu}\neq q_n)$ \\
\midrule
4  & 50.92 & 1.460 & 19.21 \\
8  & 34.93 & 0.678 & 14.93 \\
16 & 19.76 & 0.291 & 11.14 \\
32 & 8.00  & 0.101 & 6.08 \\
64 & 2.28  & 0.027 & 2.48 \\
\bottomrule
\end{tabular*}

\vspace{10pt}
\textbf{(b) Resulting symbol statistics}\par\vspace{2pt}
\setlength{\tabcolsep}{4pt}
\begin{tabular*}{295.1pt}{@{\extracolsep{\fill}}cccccc@{}}
\toprule
& \multicolumn{3}{c}{Zero-symbol ratio}
& \multicolumn{2}{c}{Relative increase} \\
\cmidrule(lr){2-4}\cmidrule(lr){5-6}
QS & No $\mu$ & Integer $\mu_n^q$ & Full $\mu$
& \shortstack{Full vs.\\No $\mu$}
& \shortstack{Full vs.\\Integer $\mu_n^q$} \\
\midrule
4  & 34.11 & 37.24 & 38.04 & 11.53 & 2.16 \\
8  & 51.85 & 55.59 & 56.73 & 9.41 & 2.05 \\
16 & 68.16 & 71.14 & 72.75 & 6.74 & 2.26 \\
32 & 84.46 & 85.99 & 87.38 & 3.45 & 1.61 \\
64 & 94.30 & 94.67 & 95.38 & 1.15 & 0.75 \\
\bottomrule
\end{tabular*}

\vspace{5pt}
\begin{tabular*}{295.1pt}{@{\extracolsep{\fill}}cccccc@{}}
\toprule
& \multicolumn{3}{c}{Mean absolute symbol value}
& \multicolumn{2}{c}{Relative reduction} \\
\cmidrule(lr){2-4}\cmidrule(lr){5-6}
QS & No $\mu$ & Integer $\mu_n^q$ & Full $\mu$
& \shortstack{Full vs.\\No $\mu$}
& \shortstack{Full vs.\\Integer $\mu_n^q$} \\
\midrule
4  & 2.885 & 2.414 & 2.398 & 16.86 & 0.66 \\
8  & 1.323 & 1.105 & 1.086 & 17.91 & 1.69 \\
16 & 0.595 & 0.504 & 0.483 & 18.90 & 4.24 \\
32 & 0.226 & 0.197 & 0.181 & 20.23 & 8.15 \\
64 & 0.072 & 0.066 & 0.058 & 19.73 & 11.85 \\
\bottomrule
\end{tabular*}

\vspace{10pt}
\textbf{(c) Coefficient reconstruction MSE}\par\vspace{2pt}
\begin{tabular*}{295.1pt}{@{\extracolsep{\fill}}cccccc@{}}
\toprule
& \multicolumn{3}{c}{Coefficient MSE}
& \multicolumn{2}{c}{MSE reduction} \\
\cmidrule(lr){2-4}\cmidrule(lr){5-6}
QS & No $\mu$ & Integer $\mu_n^q$ & Full $\mu$
& \shortstack{Full vs.\\No $\mu$}
& \shortstack{Full vs.\\Integer $\mu_n^q$} \\
\midrule
4 & 1.46 & 1.46 & 1.45 & 0.39 & 0.32 \\
8 & 6.33 & 6.30 & 6.19 & 2.20 & 1.73 \\
16 & 20.31 & 20.23 & 19.60 & 3.49 & 3.14 \\
32 & 64.98 & 64.76 & 61.44 & 5.45 & 5.14 \\
64 & 163.23 & 162.84 & 152.35 & 6.66 & 6.44 \\
\bottomrule
\end{tabular*}

\vspace{3pt}

\begin{minipage}{0.96\textwidth}
\footnotesize
Let $\mathcal{Q}_{\Delta}(r)=\mathcal{Q}(r/\Delta)$ denote the quantized
index at step $\Delta$, and let $r_0=c-\mu^{\mathrm{pre}}$ be the coefficient residual after
the preliminary prediction. The symbols without the learned correction, with
its integer component, and with the full correction are respectively
\[
q_0=\mathcal{Q}_{\Delta}(r_0),\qquad
q_n=\mathcal{Q}_{\Delta}(r_0-\mu_n),\qquad
q_{\mu}=\mathcal{Q}_{\Delta}(r_0-\mu_n-\mu_f).
\]
In (a), $\Pr(\mu_n^q\neq0)$ is the probability of a nonzero integer
correction, and $\mathbb{E}[|\mu_n^q|]$ is its mean absolute value
in quantization-index units. $\Pr(q_{\mu}\neq q_n)$ is the probability
that adding the fractional correction changes the quantized symbol.
In (b), the zero-symbol ratio is $\Pr(q=0)$,
and the mean absolute symbol value is $\mathbb{E}[|q|]$, evaluated for
$q_0$, $q_n$, and $q_{\mu}$. Panel (c) reports
$\mathbb{E}[(c-\hat c)^2]$ for the three corresponding reconstructions, where
$\hat c_0=\mu^{\mathrm{pre}}+\Delta\cdot q_0$,
$\hat c_n=\mu^{\mathrm{pre}}+\mu_n+\Delta\cdot q_n$, and
$\hat c_{\mu}=\mu^{\mathrm{pre}}+\mu_n+\mu_f+\Delta\cdot q_{\mu}$.
\end{minipage}
\end{table}

For a translation-invariant rounding quantizer, the integer component can be
absorbed exactly into the residual index. It therefore changes the coded-symbol representation without
changing reconstruction, providing an entropy-oriented residual recentering
mechanism. The fractional component cannot be absorbed into an integer symbol;
it adjusts the reconstruction within a quantization step and may also move the
residual across a quantization boundary. The actual codec uses a deadzone
offset of $1/3$, for which this separation is approximate rather than exact.

The results nevertheless exhibit the expected division of roles. Integer
correction concentrates the coded symbols closer to zero while only slightly
changing coefficient reconstruction error
(Table~\ref{tab:ford_mu_symbol_statistics}(b) and (c)), indicating that its primary
effect is residual recentering. Adding the fractional component further
increases the zero-symbol ratio and reduces reconstruction error across the
evaluated QS settings.
Although changes to the quantized symbols become less frequent at coarser QS
(Table~\ref{tab:ford_mu_symbol_statistics}(a)), reconstruction still improves,
consistent with fractional correction refining reconstructed values without
requiring a symbol change. These results support
the two practical roles of explicit prediction: residual recentering for
improved symbol concentration and sub-step reconstruction refinement for
quantization-aware distortion reduction.

These local analyses complement the rate--distortion results in
Section~\ref{sec:prediction_backend_analysis}.

\clearpage
\subsection{Complete Rate--Distortion Results}

\begin{figure}[H]
  \centering
  \begin{subfigure}[t]{0.47\textwidth}
    \centering
    \includegraphics[width=\linewidth]{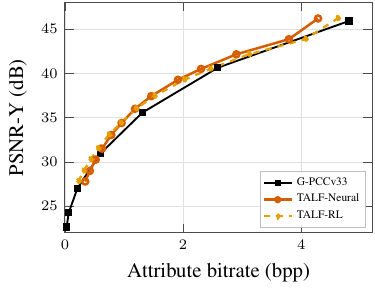}
    \caption{Luma (Y).}
    \label{fig:appendix_rd_scannet_y}
  \end{subfigure}
  \hfill
  \begin{subfigure}[t]{0.47\textwidth}
    \centering
    \includegraphics[width=\linewidth]{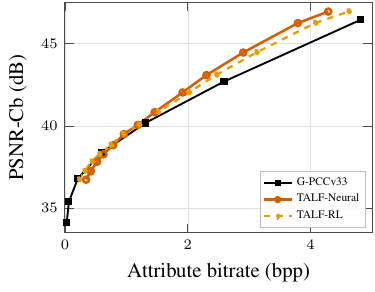}
    \caption{Chroma (Cb).}
    \label{fig:appendix_rd_scannet_cb}
  \end{subfigure}

  \vspace{10pt}

  \begin{subfigure}[t]{0.47\textwidth}
    \centering
    \includegraphics[width=\linewidth]{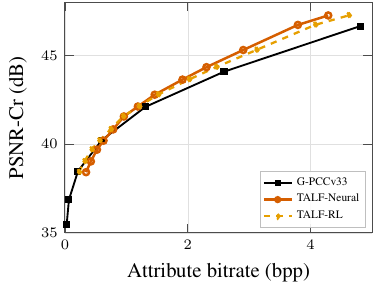}
    \caption{Chroma (Cr).}
    \label{fig:appendix_rd_scannet_cr}
  \end{subfigure}
  \caption{Rate--distortion curves for ScanNet, measured by
  (a) PSNR-Y, (b) PSNR-Cb, and (c) PSNR-Cr (dB).
  Bitrate is the total color-attribute bitrate in bits per point (bpp).}
  \label{fig:appendix_complete_rd}
\end{figure}

The complete dataset-average rate--distortion points for Ford, KITTI, and
ScanNet are provided in
Tables~\ref{tab:complete_ford_rd}--\ref{tab:complete_scannet_rd}, with the
corresponding per-channel ScanNet curves shown in
Figure~\ref{fig:appendix_complete_rd}. G-PCCv33 reports the full eight-point
TMC13 curve used for BD-Rate. All TALF points within each dataset are
produced by a single model over the same QS range used in the main paper.

\begin{table}[H]
\centering
\caption{Dataset-average rate--distortion results for Ford.}
\label{tab:complete_ford_rd}
\small
\renewcommand{\arraystretch}{1.34}
\setlength{\tabcolsep}{18pt}
\begin{tabular}{cccccc}
\toprule
\multicolumn{3}{c}{G-PCCv33} & \multicolumn{3}{c}{TALF} \\
\cmidrule(lr){1-3}\cmidrule(lr){4-6}
Point & bpp & PSNR-Refl & QS & bpp & PSNR-Refl \\
\midrule
R8 & 4.4898 & 50.17 & 4.00  & 2.9921 & 46.28 \\
R7 & 3.5089 & 46.24 & 5.03  & 2.6260 & 43.84 \\
R6 & 2.5609 & 41.00 & 6.34  & 2.3246 & 42.03 \\
R5 & 1.6631 & 35.45 & 8.00  & 2.0329 & 40.16 \\
R4 & 0.8915 & 30.23 & 10.06 & 1.8060 & 38.83 \\
R3 & 0.3508 & 25.85 & 12.69 & 1.5177 & 36.79 \\
R2 & 0.1036 & 22.84 & 16.00 & 1.2923 & 35.21 \\
R1 & 0.0297 & 21.05 & 20.13 & 1.0602 & 33.44 \\
-- & -- & -- & 25.38 & 0.8775 & 31.95 \\
-- & -- & -- & 32.00 & 0.6822 & 30.25 \\
-- & -- & -- & 40.25 & 0.5426 & 28.90 \\
-- & -- & -- & 50.75 & 0.4165 & 27.55 \\
-- & -- & -- & 64.00 & 0.3166 & 26.29 \\
\bottomrule
\end{tabular}
\end{table}

\vspace{10pt}

\begin{table}[H]
\centering
\caption{Dataset-average rate--distortion results for KITTI.}
\label{tab:complete_kitti_rd}
\small
\renewcommand{\arraystretch}{1.34}
\setlength{\tabcolsep}{18pt}
\begin{tabular}{cccccc}
\toprule
\multicolumn{3}{c}{G-PCCv33} & \multicolumn{3}{c}{TALF} \\
\cmidrule(lr){1-3}\cmidrule(lr){4-6}
Point & bpp & PSNR-Refl & QS & bpp & PSNR-Refl \\
\midrule
R8 & 3.6923 & 50.43 & 4.00  & 2.3908 & 46.44 \\
R7 & 2.7439 & 46.44 & 5.03  & 2.0212 & 43.91 \\
R6 & 1.8133 & 41.06 & 6.34  & 1.7255 & 42.07 \\
R5 & 0.8855 & 35.56 & 8.00  & 1.4498 & 40.30 \\
R4 & 0.2518 & 31.28 & 10.06 & 1.2438 & 38.97 \\
R3 & 0.0486 & 29.11 & 12.69 & 0.9629 & 37.01 \\
R2 & 0.0113 & 27.97 & 16.00 & 0.7651 & 35.57 \\
R1 & 0.0043 & 27.31 & 20.13 & 0.5682 & 34.05 \\
-- & -- & -- & 25.38 & 0.4235 & 32.88 \\
-- & -- & -- & 32.00 & 0.2702 & 31.62 \\
-- & -- & -- & 40.25 & 0.1851 & 30.79 \\
-- & -- & -- & 50.75 & 0.1267 & 30.04 \\
-- & -- & -- & 64.00 & 0.0945 & 29.45 \\
\bottomrule
\end{tabular}
\end{table}

\begin{table}[H]
\centering
\caption{Dataset-average rate--distortion results for ScanNet.}
\label{tab:complete_scannet_rd}
\small
\renewcommand{\arraystretch}{1.14}
\setlength{\tabcolsep}{2pt}
\begin{tabular*}{385.775pt}{@{\extracolsep{\fill}}ccccccccccccc@{}}
\toprule
\multicolumn{6}{c}{G-PCCv33} & \multicolumn{7}{c}{TALF} \\
\cmidrule(lr){1-6}\cmidrule(lr){7-13}
Point & bpp & Y & Cb & Cr & YCbCr & QS & Neural bpp & RL bpp & Y & Cb & Cr & YCbCr \\
\midrule
R8 & 7.6590 & 49.98 & 49.98 & 49.73 & 49.95 & 4.00  & 4.2864 & 4.6183 & 46.20 & 46.95 & 47.27 & 46.41 \\
R7 & 4.8095 & 45.92 & 46.39 & 46.62 & 46.06 & 5.03  & 3.7922 & 4.0720 & 43.86 & 46.25 & 46.74 & 44.39 \\
R6 & 2.5896 & 40.63 & 42.42 & 43.99 & 41.14 & 6.34  & 2.9026 & 3.1165 & 42.16 & 44.46 & 45.32 & 42.69 \\
R5 & 1.3095 & 35.49 & 39.82 & 41.97 & 36.33 & 8.00  & 2.3043 & 2.4613 & 40.51 & 43.09 & 44.35 & 41.11 \\
R4 & 0.5993 & 30.82 & 38.06 & 40.03 & 31.85 & 10.06 & 1.9140 & 2.0176 & 39.28 & 42.03 & 43.65 & 39.92 \\
R3 & 0.2118 & 26.84 & 36.48 & 38.21 & 27.96 & 12.69 & 1.4612 & 1.5137 & 37.43 & 40.84 & 42.80 & 38.17 \\
R2 & 0.0646 & 24.04 & 35.05 & 36.60 & 25.19 & 16.00 & 1.1872 & 1.2057 & 36.01 & 40.06 & 42.13 & 36.82 \\
R1 & 0.0277 & 22.42 & 33.71 & 35.16 & 23.58 & 20.13 & 0.9591 & 0.9463 & 34.40 & 39.50 & 41.57 & 35.30 \\
-- & -- & -- & -- & -- & -- & 25.38 & 0.7849 & 0.7428 & 33.06 & 38.83 & 40.84 & 34.00 \\
-- & -- & -- & -- & -- & -- & 32.00 & 0.6275 & 0.5581 & 31.50 & 38.28 & 40.20 & 32.51 \\
-- & -- & -- & -- & -- & -- & 40.25 & 0.5208 & 0.4323 & 30.27 & 37.82 & 39.69 & 31.31 \\
-- & -- & -- & -- & -- & -- & 50.75 & 0.4217 & 0.3165 & 28.99 & 37.25 & 39.03 & 30.06 \\
-- & -- & -- & -- & -- & -- & 64.00 & 0.3435 & 0.2267 & 27.80 & 36.73 & 38.43 & 28.89 \\
\bottomrule
\end{tabular*}
\end{table}

\end{document}